\documentclass[10pt,conference]{IEEEtran}
\IEEEoverridecommandlockouts  
\usepackage[T1]{fontenc}
\usepackage{graphicx}
\usepackage[utf8]{inputenc}
\usepackage{amssymb,amsmath,array}
\usepackage{tikz}

\usepackage[utf8]{inputenc}
\usepackage[T1]{fontenc}
\usepackage{graphicx}
\usepackage{amsmath,amssymb}
\usepackage{cite}
\usepackage{url}
\usepackage{hyperref} 
\usepackage{xcolor}
\usepackage{float}
\usepackage{multirow}
\usepackage{booktabs}
\usepackage{makecell}

\title{HydraCIL: Decoupled Class-Incremental Learning through Prototype-Guided Multi-Head Classifiers}

\author{  
\IEEEauthorblockN{Daniel Vila-Cruz}  
\IEEEauthorblockA{  
\textit{Universidade da Coruña}\\  
\textit{CITIC} \\  
A Coruña, Spain \\  
0009-0001-2116-4085}  
\and  
\IEEEauthorblockN{Laura Morán-Fernández}  
\IEEEauthorblockA{  
\textit{Universidade da Coruña}\\  
\textit{CITIC} \\  
A Coruña, Spain \\  
0000-0001-6703-1846}  
\and  
\IEEEauthorblockN{Verónica Bolón-Canedo}  
\IEEEauthorblockA{  
\textit{Universidade da Coruña}\\  
\textit{CITIC} \\  
A Coruña, Spain \\  
0000-0002-0524-6427}  
\thanks{  
* Accepted for publication at the International Joint Conference on Neural Networks (IJCNN 2026).
This work was supported by the Ministry of Science and Innovation of Spain (Grant PID2023-147404OB-I00 / AEI / 10.13039 / 501100011033) and together with ``NextGenerationEU''/PRTR by the Ministry for Digital Transformation and Civil Service under grant TSI-100925-2023-1 and by Xunta de Galicia (Grants ED431G 2023/01 and ED431C 2022/44)}  
}

\begin{document}
\maketitle
\thispagestyle{plain}
\pagestyle{plain}

\begin{abstract}

We present HydraCIL, a decoupled continual learning model based on prototype-guided multi-head classifiers, targeting sustainable deployment in embedded and resource-constrained environments. While most Class-Incremental Learning (CIL) methods rely on powerful hardware and long retraining cycles, real-world systems, such as robots or edge AI devices, must adapt quickly with limited resources. HydraCIL addresses this gap by freezing the backbone and decoupling feature extraction from learning. For each task, features are extracted once and a lightweight, task-specific classifier head is created, avoiding costly backbone retraining. At inference, HydraCIL selects the appropriate head via similarity with prototypes. Experiments on CIFAR-100, ImageNet-100, CoRe50, and Flowers102 datasets show that HydraCIL matches or outperforms state-of-the-art CIL methods while significantly reducing training time and carbon footprint, making it a practical solution for continual learning in real-world and embedded settings, where energy efficiency and rapid adaptation are critical.


\end{abstract}

\section{Introduction}

Continual Learning (CL) aims to enable AI systems to learn sequentially from a stream of tasks without forgetting previously acquired knowledge. Among the various CL scenarios, Class-Incremental Learning (CIL) represents one of the most challenging settings. In CIL, new classes are introduced over time, and the model must be able to recognize all classes encountered so far, without access to the entire past data and without task identifiers at test time. Traditional neural networks, when trained sequentially on different class sets, tend to forget previously learned information, as gradient updates for new classes interfere destructively with earlier representations. To mitigate this, prior work has explored several strategies, that could be grouped into the following taxonomy, proposed in \cite{zhou2024class}: data replay, dynamic networks, parameter regularization, knowledge distillation, model rectify, weight rectification and template-based classification. However, existing solutions often face a difficult trade-off between accuracy, scalability and computational cost. In practical applications like autonomous robotics or edge AI, models must adapt rapidly to new data while operating under strict constraints on memory, energy consumption, and training latency. Most state-of-the-art CIL methods assume powerful hardware and long training cycles, rarely reporting environmental metrics like energy usage or CO2 emissions. This creates a critical gap between theoretical research and the deployment of sustainable AI systems.

These constraints have a direct impact on architectural choices in continual learning. While recent advances in the field increasingly rely on Vision Transformers (ViTs) due to their strong representational capacity, such architectures are often prohibitively large for embedded and edge devices. There are multiple techniques aiming to compress and accelerate transformer-based models, but adapting these models to specific hardware requirements is still a costly and manual labor \cite{saha2025vision}. As a result, Convolutional Neural Networks (CNNs) remain the dominant choice in many real-world applications.

In this work, we introduce HydraCIL, a decoupled class-incremental learning system based on prototype-guided multi-head classifiers, explicitly designed for embedded and resource-aware continual learning. HydraCIL spawns lightweight classifier heads for each incremental task while keeping the backbone frozen, drastically reducing training time and energy consumption. To enable task-agnostic inference, we employ a similarity-based routing mechanism using class prototypes, allowing the model to dynamically select the most relevant classifier head at inference time.

The remainder of this paper is organized as follows. Section \ref{Sec:state-of-the-art} reviews the state-of-the-art of CIL methods. Section \ref{Sec:methodology} describes the proposed methodology. Section \ref{Sec:evaluation} explains the evaluation setup. Section \ref{Sec:results} details the experimental results. Finally, Section \ref{Sec:conclusions} summarizes the conclusions of this work.

\section{State-of-the-art}
\label{Sec:state-of-the-art}


Class-Incremental Learning has seen a surge in interest as researchers seek to overcome catastrophic forgetting. Current strategies to mitigate this include data replay, dynamic networks, knowledge distillation, and parameter regularization. However, these approaches often face a difficult trade-off between accuracy, scalability, and computational cost. This is particularly problematic for real-world deployment in embedded and edge devices, where strict constraints on memory, energy consumption, and training latency make many high-performance, transformer-based models prohibitively expensive.

Dynamic network methods address catastrophic forgetting by expanding the model architecture over time, allocating new parameters to incoming tasks or classes while preserving previously learned knowledge. Representative methods include DER \cite{yan2021dynamically}, FOSTER \cite{wang2022foster}, and MEMO \cite{zhou2022model}, which dynamically grow feature representations or classifier components to reduce interference between old and new classes. These approaches have demonstrated strong performance in challenging CIL benchmarks by dynamically increasing model capacity.

However, most dynamic network methods rely on joint optimization of the backbone and newly added components, often in combination with replay or distillation mechanisms. As a result, training costs tend to grow significantly with the number of tasks, leading to long training times and high energy consumption. Moreover, architectural growth typically affects inference complexity or memory usage, which can be problematic in embedded or resource-constrained environments.


Data replay approaches mitigate forgetting by storing or generating samples from previously learned classes and replaying them during the training of new tasks. Methods such as DER \cite{yan2021dynamically} combine replay with dynamically expandable representations, while RMM \cite{liu2021rmm} integrates memory management strategies. 

Despite their effectiveness, replay methods require additional memory and computational overhead, as old samples must be processed repeatedly through the backbone. This makes them less suitable for continual learning in embedded systems or long-term deployments, where memory buffers and repeated backbone updates are costly. In contrast, there is a family of class-incremental learning that avoids the replay of raw samples from previous tasks, that is Exemplar-Free Class-Incremental Learning (EFCIL). Approaches such as SSRE \cite{zhu2022self}, FeTrIL \cite{petit2023fetril} and RRFE \cite{luo2023representation} focus on improving representation robustness or expanding feature spaces to preserve class separability without explicit replay, through the use of prototypes. 


Another relevant distinction among CIL methods lies in their task initialization assumptions. In this work, we adopt the $B=0$ setting, where all incremental tasks contain the same number of classes, ensuring a uniform difficulty across the learning sequence and avoiding privileged access to a larger initial dataset. In contrast, several recent methods, including FeTrIL \cite{petit2023fetril} and SSRE \cite{zhu2022self}, rely on a larger first task to adequately train or adapt the backbone representation before incremental updates. While this design choice can improve performance, it introduces an implicit advantage that may not be feasible in real-world continual learning scenarios, particularly in embedded or deployed systems where data availability is limited and class distributions are not known in advance.

Efficiency on CIL methods is already a concern. As noted in \cite{harun2023efficient}, many state-of-the-art approaches require more computational power than the equivalent offline learner, which makes them irrelevant for many industrial applications. In this study, they found that REMIND \cite{hayes2020remind} achieves the best efficiency score, but accuracy falls compared to other methods like DER. Some models emerged focused on avoiding catastrophic forgetting while reducing training costs, like CIFNet \cite{dopico2025efficient}, that also relies on a pre-trained backbone.

HydraCIL follows a dynamic network paradigm, but differs fundamentally from existing approaches by freezing the backbone and restricting architectural expansion to lightweight task-specific classifier heads. This design limits computational growth while preserving the benefits of dynamic adaptation. Additionally, it generates prototypes of data, but never uses them to train the model, like other data replay methods.

\section{Proposed methodology}
\label{Sec:methodology}
We propose a three-step train and three-step inference model designed to maximize classification accuracy while minimizing computational time and energy consumption. The key architectural design that allows the model to reduce training times is to decouple the feature extractor (backbone) from the classifiers (heads). Most fine-tuning approaches rely on freezing most layers of the backbone and train the remaining ones with the head. Freezing the whole backbone allows to process all data that belongs to the current task just once, avoiding multiple operations with the same deterministic features. 

\subsection{Training}




The training phase is organized into three sequential steps: feature extraction, classifier-head training, and prototype generation. When a new task arrives, the frozen backbone processes all input samples only once, producing fixed feature representations that are reused in the remaining stages. In practice, the backbone encodes each input into a lower-dimensional vector, for example, a 224×224 RGB image (150,528 pixel values) is compressed into a 512-dimensional feature representation. This substantially reduces memory usage and allows larger batch sizes during head training. After the feature extraction step, the model spawns a new classifier head, that is trained using only the corresponding task features. Since the head is specialized only in the task features, typically a small fraction of the overall problem, a complex architecture is not required. Additionally, the model creates a new head for each task, so keeping its architecture simple reduces the memory requirements. This design choice is computationally efficient, as the inference process (explained in the next subsection) does not require all heads to be active simultaneously. Therefore, if the model requires a more complex classifier architecture due to the task split or the nature of the data, the inference time and memory requirements do not grow linearly. Once the head is trained, $k$ prototypes are created for each class. These prototypes are later used to determine which classifier head should be selected during inference. To generate these prototypes, we apply $k$-means clustering, where $k$ is a hyperparameter that specifies the number of prototypes per class. The final prototypes correspond to the centroids of the resulting clusters. Using multiple prototypes per class, rather than a single class centroid, allows the model to account for intra-class variability.


\subsection{Inference}

The inference stage also follows three steps: feature extraction, classifier-head selection via prototype comparison, and prediction. The feature extraction process is identical to that used during training, as the backbone remains frozen and produces deterministic representations for each input. Once the features are obtained, they are compared against the prototypes of all classes. Using the Euclidean distance, the model selects the classifier head whose prototypes best match the input representation, and this head is then used to produce the final prediction. The idea is that each head is specialized in a different task. Since there is no guarantee that tasks are semantically related, two tasks can contain very similar classes.  Consequently, a sample may fall close to prototypes from two different tasks if their classes are semantically similar. In this case, the system allows the classifiers of both tasks to perform inference on the sample, and the prediction with the highest confidence score is selected as the final output. This approach ensures that the system only uses the required classifiers, instead of keeping all heads active in memory or running inference with all of them simultaneously. Since prototypes are class-related and not task-related, it is also possible to use them directly to make the predictions, instead of using an additional classifier. This was also tested, but accuracy dropped noticeably.

\section{Evaluation}
\label{Sec:evaluation}





\subsection{Metrics}
The metrics employed to perform the evaluation are the following ones:
\begin{itemize}
    \item \textbf{Final classification accuracy (\%)}: Measured once the entire sequence of incremental tasks has been presented to the model. This represents the system’s ability to recognize all encountered classes without the aid of task identifiers. 
    
    \item \textbf{Training time (minutes)}: This metric encompasses the total time required to adapt the model to new data and validate it. For HydraCIL, this includes the entire pipeline: the single-pass feature extraction through the frozen backbone, the gradient updates for the task-specific heads, the generation of class prototypes, and the head selection and classification of the current task data, performed at validation time.
 
    \item \textbf{Energy consumption (kWh)}: Recorded via the Codecarbon library \cite{benoit_courty_2024_11171501}, this represents the electrical cost of the adaptation process. 
    
    \item \textbf{Carbon emissions (kg CO2-eq)}: Also tracked through Codecarbon, this quantifies the environmental footprint of the training sequence. For doing so, it takes into account the carbon intensity of the local energy grid, providing a real-world estimate of the environmental cost associated. 
    
\end{itemize}

\subsection{Experimental configurations}
We evaluated HydraCIL using three configurations for the number of prototypes per class: $k \in \{2, 5, 20\}$. To ensure a fair comparison with replay-based methods that typically utilize a fixed memory buffer of 2,000 samples, our $k=20$ setting was chosen to mirror this storage overhead. Specifically, for our 100-class benchmarks (CIFAR-100 and ImageNet-100), $k=20$ corresponds exactly to the 20 samples-per-class allocation used by the baselines. This equivalency holds across our other benchmarks: for Flowers102 (102 classes), $k=20$ results in 2,040 prototypes, while for CoRe50 (50 classes), it results in 1,000 prototypes. It is important to note that while the count is equivalent, our prototypes consist of condensed feature vectors ($1 \times 512$ for ResNet-34) rather than raw images, and they are utilized exclusively during the inference-time head selection process.

All the models that require the use of a backbone were tested using ResNet-34, and the ones that use a pretrained version, employ ResNet-34 pretrained on ImageNet-1k \cite{imagenet}. Experiments were conducted using an NVIDIA GeForce RTX 3090 Ti GPU and an Intel Core i7-12700K CPU. Tasks were randomly split across classes to ensure semantic diversity among tasks (i.e., similar classes could appear in different tasks).

\subsection{Comparative baselines}
We compared HydraCIL with the top-3 performing models reported in \cite{dopico2025efficient}: RMM-FOSTER, which combines RMM \cite{liu2021rmm} (\textit{data replay})  and FOSTER \cite{wang2022foster} (\textit{dynamic network}), DER \cite{yan2021dynamically} (\textit{dynamic network}) and CIFNet \cite{dopico2025efficient} (\textit{data replay}). Additionally, among the prototype-based models, FeTrIL was also compared, as it has the faster training times, making it more suitable to our case of use. This model requires a first phase to train the backbone, using a significant amount of classes, so its task distribution configuration differs from other models. A configuration of $B=0$ was adopted for all but this model, meaning that all incremental steps contain an equal number of classes, ensuring uniform difficulty progression throughout the sequence. Additionally, our approach uses a pre-trained backbone, so it performs better by resizing data to the size it was trained on, 224x224 in this case. This is a limitation that FeTrIL does not have, so we tested it using original data size and upscaled data. We found that it provided better accuracy using original data sizes, so, trying to be as fair as possible, and as this is a limitation of our model, we report metrics of FeTrIL using original data sizes.

\subsection{Benchmarks}

To validate the proposed approach, we evaluated HydraCIL on four widely used benchmarks for Class-Incremental Learning: CIFAR-100 \cite{cifar100}, ImageNet-100 \cite{imagenet}, CoRe50 \cite{lomonaco2017core50} and Flowers102 \cite{nilsback2008automated}.

\subsubsection{CIFAR-100}
This dataset consists of 100 classes, comprising 50,000 training images and 10,000 test images. To align with the input requirements of the pre-trained ResNet-34 backbone, images are upscaled to 224x224 pixels for all models except FeTrIL, which utilize their native 32x32 resolution to maintain their reported performance standards. We evaluate this benchmark using two primary configurations: a 20-task sequence (5 classes per task) and a 5-task sequence (20 classes per task). Following the protocol established in prior work, the FeTrIL baseline is evaluated using a $B=50$ setup, where 50 classes are used for initial backbone adaptation followed by 10 incremental tasks of 5 classes each.

\subsubsection{ImageNet-100} \label{imagenet_benchmark}
This dataset is a subset of the ImageNet-1k dataset containing 100 classes with approximately 130,000 training and 5,000 test images. All images are utilized at 224x224 resolution. We evaluated performance using two incremental schedules: a 20-task sequence (5 classes per task) and a 10-task sequence (10 classes per task). For the FeTrIL baseline, we adopt a $B=50$ configuration followed by 10 tasks of 5 classes each. 

It is important to acknowledge that both HydraCIL and CIFNet utilize a backbone that was pre-trained on ImageNet-1k, so these models have an advantage on this dataset, as their feature extractor has some prior knowledge on the data. Even knowing this, we considered adding the dataset for evaluation as it can bring more information regarding training times, energy consumption and CO2 emissions when adapting the model to larger datasets.

\subsubsection{CoRe50}
This dataset is specifically designed for continuous object recognition in robotics, consisting of 50 domestic objects across 10 categories. The dataset comprises 11 distinct sessions with varying backgrounds and lighting conditions. To mitigate data redundancy inherent in the original 15-second video recordings (approximately 164,866 images), we performed downsampling by selecting every $10^{th}$ frame. This resulted in a more manageable and diverse subset of 16,486 images. Images were resized to 224x224 for all models except FeTrIL, which utilized the original 128x128 resolution. We evaluate the model at the object level (50 classes) across two settings: 10 tasks (5 classes per task) and 5 tasks (10 classes per task). The FeTrIL configuration utilizes $B=25$ with 5 subsequent increments of 5 classes.

\subsubsection{Flowers102}
This dataset consist of 102 flower categories. A notable challenge of this benchmark is the class imbalance, with sample counts ranging from 40 to 258 per category. The total 8,200 images were resized to 224x224 for all models. Given the limited sample sizes for certain classes, we focused on one specific configuration: a 6-task sequence (17 classes per task). The FeTrIL configuration utilizes $B=52$ with 5 subsequent increments of 10 classes.



\section{Experimental results}
\label{Sec:results}




Tables \ref{Tab:cifar100-20t} and \ref{Tab:cifar100-5t} summarize the results for CIFAR-100, using 20 tasks and 5 tasks, respectively. Our model achieves higher accuracy than all other methods when splitting the data into 20 tasks, regardless of the chosen value of $k$. When reducing the setup to only 5 tasks, RMM-FOSTER and DER slightly outperform our model for $k=2$, although HydraCIL remains the best performing approach when using $k=20$. Overall, all models achieve higher accuracy with fewer tasks, but our approach benefits less from this, as it is designed to perform better on small tasks. Since our model relies on very simple classifiers to separate few classes, its performance decreases on more complex problems. This could be adapted by increasing classifiers complexity, adding minimal computation overhead, as proposed in other decoupling frameworks \cite{vila2025fast}. 
Consequently, while two models outperform ours using $k=2$ on 5 tasks, their accuracy drops significantly as the number of tasks increases, whereas our model shows a more gradual decrease. Regarding training times, RMM-FOSTER and DER scale poorly as the number of tasks increases, while CIFNet and HydraCIL remain with similar times. Moreover, HydraCIL achieves a substantial reduction of training times compared to other approaches. Considering the configuration that surpasses all other methods in accuracy, for 20 tasks and $k=2$, HydraCIL is 54x faster than RMM-FOSTER, 73x faster than DER, and 11x faster than CIFNet. In the 5-task setting, and using $k=20$, our model is 24x faster than RMM-FOSTER, 18x faster than DER, and 9x faster than CIFNet.



\begin{table}[htb!]
    \setlength{\tabcolsep}{4pt}
  \centering
  \caption{Results of CIFAR-100 with 20 tasks, 5 classes per task. Best results are highlighted in bold.}\label{Tab:cifar100-20t}
  \begin{tabular}{c|c|c|c|c}
    \hline
    Model & Accuracy & \begin{tabular}{@{}c@{}}Time \\ (min)\end{tabular} & \begin{tabular}{@{}c@{}}Energy \\  (kWh)\end{tabular} & \begin{tabular}{@{}c@{}}Emissions \\ (kg CO2-eq)\end{tabular} \\
    \hline
    Fine-tune & 5.25 & 14.17 & 0.068 & 0.012 \\
    \hline
    RMM-FOSTER & 51.10 & 76.36 & 0.429 & 0.075\\
    DER & 57.34 & 102.44 & 0.667 & 0.116\\
    CIFNet& 59.26 & 15.13 & 0.063 & 0.011 \\
    \hline
    HydraCIL, $k=2$  & 60.73 & \textbf{1.41} & \textbf{0.005} & $\mathbf{8.4 \times 10^{-4}}$ \\
    HydraCIL, $k=5$  & 62.01 & 1.56 & \textbf{0.005} & $9.5 \times 10^{-4}$ \\
    HydraCIL, $k=20$ & \textbf{62.66} & 2.02 & 0.007 & 0.001 \\
    \hline
  \end{tabular}
  
\end{table}

\begin{table}[htb!]
    \setlength{\tabcolsep}{4pt}
  \centering
  \caption{Results of CIFAR-100 with 5 tasks, 20 classes per task. Best results are highlighted in bold.}\label{Tab:cifar100-5t}
  \begin{tabular}{c|c|c|c|c}
    \hline
    Model & Accuracy & \begin{tabular}{@{}c@{}}Time \\ (min)\end{tabular} & \begin{tabular}{@{}c@{}}Energy \\  (kWh)\end{tabular} & \begin{tabular}{@{}c@{}}Emissions \\ (kg CO2-eq)\end{tabular} \\
    \hline
    Fine-tune & 17.14 & 10.91 & 0.061 & 0.011 \\
    \hline
    RMM-FOSTER & 63.01 & 37.55 & 0.227 & 0.041 \\
    DER & 62.67 & 28.50 & 0.172 & 0.032\\
    CIFNet& 59.98 & 13.90 & 0.058  & 0.010 \\
    
    \hline
    HydraCIL, $k=2$  & 61.86 & \textbf{1.30} & \textbf{0.004} & $\mathbf{7.8 \times 10^{-4}}$ \\
    HydraCIL, $k=5$ & 62.75 & 1.37 & 0.005 & $8.1 \times 10^{-4}$ \\
    HydraCIL, $k=20$  & \textbf{63.87} & 1.57 & 0.005 & $9.2 \times 10^{-4}$ \\
    \hline
  \end{tabular}
  
\end{table}

The results for the ImageNet-100 benchmark, presented in tables \ref{Tab:imagenet100-20t} and \ref{Tab:imagenet100-10t}, represent the results for ImageNet-100, further validate the scalability of HydraCIL. Across both 20-task and 10-task settings, our model achieves the highest accuracy while maintaining a negligible computational footprint. As mentioned in Section \ref{imagenet_benchmark}, our backbone was pre-trained on ImageNet-1k, so accuracy comparison is only fair with CIFNet, as it uses the same pre-trained backbone. Our proposal ourperforms CIFNet by over 8 percentage points (86.26\% vs 78.10\%) in the 20-task setting. The most valuable result relies on computational metrics. In the 20-task experiment, HydraCIL ($k=5$) completes the entire incremental sequence in just 6.94 minutes. Compared to traditional dynamic networks like DER (1354.40 min) and RMM-FOSTER (935.67 min), HydraCIL provides a staggering 195$\times$ and 135$\times$ speedup, respectively. Even against the efficient CIFNet baseline, our approach is 10$\times$ faster. The energy requirements of traditional approaches are substantial. DER required 9.972 kWh to be trained, equivalent to the carbon footprint of approximately 1.736 kg CO2-eq. HydraCIL reduces this to a mere 0.020 kWh (0.003 kg CO2-eq).


\begin{table}[htb]
    \setlength{\tabcolsep}{4pt}
  \centering
  \caption{Results of ImageNet-100 with 20 tasks, 5 classes per task. Best results are highlighted in bold.}\label{Tab:imagenet100-20t}
  \begin{tabular}{c|c|c|c|c}
    \hline
    Model & Accuracy & \begin{tabular}{@{}c@{}}Time \\ (min)\end{tabular} & \begin{tabular}{@{}c@{}}Energy \\ (kWh)\end{tabular} & \begin{tabular}{@{}c@{}}Emissions \\ (kg CO2-eq)\end{tabular} \\
    \hline
    Fine-tune & 4.64 & 188.93 & 1.136 & 0.198 \\
    \hline
    RMM-FOSTER & 59.46 & 935.67 & 6.255 & 1.089\\
    DER & 63.66 & 1354.40 & 9.972 & 1.736\\
    CIFNet& 78.10 & 71.50 & 0.271 & 0.047 \\
    \hline
    HydraCIL $k=2$  & 85.72 & 7.32 & 0.021 & 0.004\\
    HydraCIL $k=5$ & 85.48 & \textbf{6.94} & \textbf{0.020} & \textbf{0.003}\\
    HydraCIL $k=20$  & \textbf{86.26} & 7.29 & 0.021 & 0.004\\
    \hline
  \end{tabular}
  
\end{table}

\begin{table}[htb]
    \setlength{\tabcolsep}{4pt}
  \centering
  \caption{Results of ImageNet100 with 10 tasks, 10 classes per task. Best results are highlighted in bold.}\label{Tab:imagenet100-10t}
  \begin{tabular}{c|c|c|c|c}
    \hline
    Model & Accuracy & \begin{tabular}{@{}c@{}}Time \\ (min)\end{tabular} & \begin{tabular}{@{}c@{}}Energy \\  (kWh)\end{tabular} & \begin{tabular}{@{}c@{}}Emissions \\ (kg CO2-eq)\end{tabular} \\
    \hline
    Fine-tune & 9.34 & 184.20 & 1.152 & 0.201  \\
    \hline
    RMM-FOSTER & 66.10 & 757.55 & 5.186 & 0.903 \\
    DER & 64.40 & 742.22 & 5.328 & 0.927 \\
    CIFNet& 77.92 & 70.02 & 0.264 & 0.046  \\
    \hline
    HydraCIL $k=2$  & 85.66 & 6.72 & \textbf{0.019} & \textbf{0.003} \\
    HydraCIL $k=5$  & 85.50 & \textbf{6.68} & 0.020 & \textbf{0.003} \\
    HydraCIL $k=20$  & \textbf{85.70} & 7.11 & 0.021 & 0.004 \\
    \hline
  \end{tabular}
  
\end{table}

Tables \ref{Tab:core50-incr5} and \ref{Tab:core50-incr10} contain the results of experiments with CoRe50, using 10 and 5 tasks, respectively. For the first one, HydraCIL $k=5$ achieves best accuracy (72.29\%), very close to CIFNet (72.22\%), that surpasses HydraCIL using other $k$ settings. Other methods remain far away regarding accuracy, being the third best one DER (51.33\%), with more than 20\% of difference. In the 10-task experiment, HydraCIL completes the training in roughly 0.59 minutes, compared to 401.45 minutes for RMM-FOSTER and 295.55 minutes for DER. This represents a speedup of 500$\times$ compared to DER and a speedup of over 680$\times$ compared to RMM-FOSTER. The energy consumption of 0.002 kWh suggest that HydraCIL could feasibly perform continuous learning updates on a mobile robot’s embedded hardware with minimal impact on battery life.

\begin{table}[htb]
    \setlength{\tabcolsep}{4pt}
  \centering
  \caption{Results of CORE50 with 10 tasks, 5 classes per task. Best results are highlighted in bold.}
  \label{Tab:core50-incr5}
  \begin{tabular}{c|c|c|c|c}
    \hline
    Model & Accuracy & \begin{tabular}{@{}c@{}}Training time \\ (min)\end{tabular} & \begin{tabular}{@{}c@{}}Energy \\ (kWh)\end{tabular} & \begin{tabular}{@{}c@{}}Emissions \\ (kg CO2-eq)\end{tabular} \\
    \hline
    Fine-tune & 19.11 & 35.64 & 0.186 & 0.032 \\
    \hline
    RMM-FOSTER & 38.82 & 401.45 & 2.402 & 0.418 \\
    DER        & 51.33 & 295.55 & 1.787 & 0.311 \\
    CIFNet     & 72.22 & 3.63 & 0.013 & 0.002 \\
    \hline
    HydraCIL $k=2$  & 72.20 & \textbf{0.58} & \textbf{0.002} & $\mathbf{4.1 \times 10^{-4}}$ \\
    HydraCIL $k=5$  & \textbf{72.29} & 0.59 & \textbf{0.002} & $\mathbf{4.1 \times 10^{-4}}$  \\
    HydraCIL $k=20$ & 72.04 & 0.61 & \textbf{0.002} & $4.2 \times 10^{-4}$ \\
    \hline
  \end{tabular}
\end{table}

\begin{table}[htb]
    \setlength{\tabcolsep}{4pt}
  \centering
  \caption{Results of CORE50 with 5 tasks,  10 classes per task. Best results are highlighted in bold.}
  \label{Tab:core50-incr10}
  \begin{tabular}{c|c|c|c|c}
    \hline
    Model & Accuracy & \begin{tabular}{@{}c@{}}Time \\ (min)\end{tabular} & \begin{tabular}{@{}c@{}}Energy \\ (kWh)\end{tabular} & \begin{tabular}{@{}c@{}}Emissions \\ (kg CO2-eq)\end{tabular} \\
    \hline
    Fine-tune & 19.11 & 31.61 & 0.186 & 0.032 \\
    \hline
    RMM-FOSTER & 48.76 & 255.92 & 1.544 & 0.269 \\
    DER        & 47.61 & 134.76 & 0.804 & 0.140 \\
    CIFNet & 72.33 & 3.21 & 0.002 & 0.011 \\
    \hline
    HydraCIL $k=2$  & \textbf{72.96} & \textbf{0.56} & \textbf{0.002} & $\mathbf{3.8 \times 10^{-4}}$ \\
    HydraCIL $k=5$  & 72.24 & \textbf{0.56} & \textbf{0.002} & $4.0 \times 10^{-4}$ \\
    HydraCIL $k=20$ & 72.29 & 0.60 & \textbf{0.002} & $4.1 \times 10^{-4}$ \\
    \hline
  \end{tabular}
\end{table}

The results for Flowers102, detailed in Table \ref{Tab:flowers102}, showcase HydraCIL’s efficacy in fine-grained classification tasks. In this 6-task scenario, HydraCIL ($k=2$) achieves a peak accuracy of 74.99\%, surpassing the strongest baseline, CIFNet, by 4.06 percentage points. Notably, the traditional dynamic network approaches, DER (22.89\%) and RMM-FOSTER (35.71\%), struggle significantly with this dataset, likely due to the difficulty of distinguishing between similar flower species. An interesting observation in this benchmark is the inverse relationship between the number of prototypes ($k$) and accuracy. Unlike the previous datasets where higher $k$ often yielded better results, here $k=2$ outperforms $k=20$ by over 7\%. This suggests that in domains where class boundaries are narrow, a higher density of prototypes may lead to overlapping decision regions in the feature space, causing the model to misidentify the correct task head.

\begin{table}[htb]
    \setlength{\tabcolsep}{4pt}
  \centering
  \caption{Results of Flowers102 with 6 tasks, 17 classes per task. Best results are highlighted in bold.}
  \label{Tab:flowers102}
  \begin{tabular}{c|c|c|c|c}
    \hline
    Model & Accuracy & \begin{tabular}{@{}c@{}}Time \\ (min)\end{tabular} & \begin{tabular}{@{}c@{}}Energy \\ (kWh)\end{tabular} & \begin{tabular}{@{}c@{}}Emissions \\ (kg CO2-eq)\end{tabular} \\
    \hline
    Fine-tune & 20.99 & 15.91 & 0.070 & 0.012 \\
    \hline
    RMM-FOSTER & 35.71 & 69.63 & 0.337 & 0.068 \\
    DER        & 22.89 & 59.81 & 0.321 & 0.065 \\
    CIFNet     & 70.93 & 1.41 & 0.004 & $7.7 \times 10^{-4}$ \\
    \hline
    HydraCIL $k=2$  & \textbf{74.99} & \textbf{0.45} & \textbf{0.001} & $\mathbf{1.7 \times 10^{-4}}$ \\
    HydraCIL $k=5$  & 70.94 & 0.46 & \textbf{0.001} & $\mathbf{1.7 \times 10^{-4}}$ \\
    HydraCIL $k=20$ & 67.31 & \textbf{0.45} & \textbf{0.001} & $\mathbf{1.7 \times 10^{-4}}$ \\
    \hline
  \end{tabular}
\end{table}

Table \ref{Tab:fetril} provides an overview of FeTrIL's performance across all four benchmarks. Despite the advantages provided to FeTrIL, the use of a large base task ($B \in \{25, 50, 52\}$) and optimal original image resolutions, it performs worse than HydraCIL and CIFNet on all metrics. 

\begin{table}[htb!]
    \setlength{\tabcolsep}{4pt}
  \centering
  \caption{Results of FeTrIL evaluation among different datasets.
}\label{Tab:fetril}
  \begin{tabular}{c|c|c|c|c}
    \hline
    Dataset & Accuracy & \begin{tabular}{@{}c@{}}Time \\ (min)\end{tabular} & \begin{tabular}{@{}c@{}}Energy \\  (kWh)\end{tabular} & \begin{tabular}{@{}c@{}}Emissions \\ (kg CO2-eq)\end{tabular} \\
    \hline
    CIFAR-100 ($B=50$) & 53.31 & 22.55 & 0.062 & 0.010 \\
    ImageNet-100 ($B=50$) & 71.68 & 295.65 & 1.672 & 0.291 \\
    CoRe50 ($B=25$) & 38.12 & 14.17 & 0.075 & 0.013 \\
    Flowers102 ($B=52$) & 25.52 & 19.33 & 0.077 & 0.014 \\
    \hline
  \end{tabular}
  
\end{table}

\begin{figure}[htb]
\centering
\includegraphics[scale=0.22]{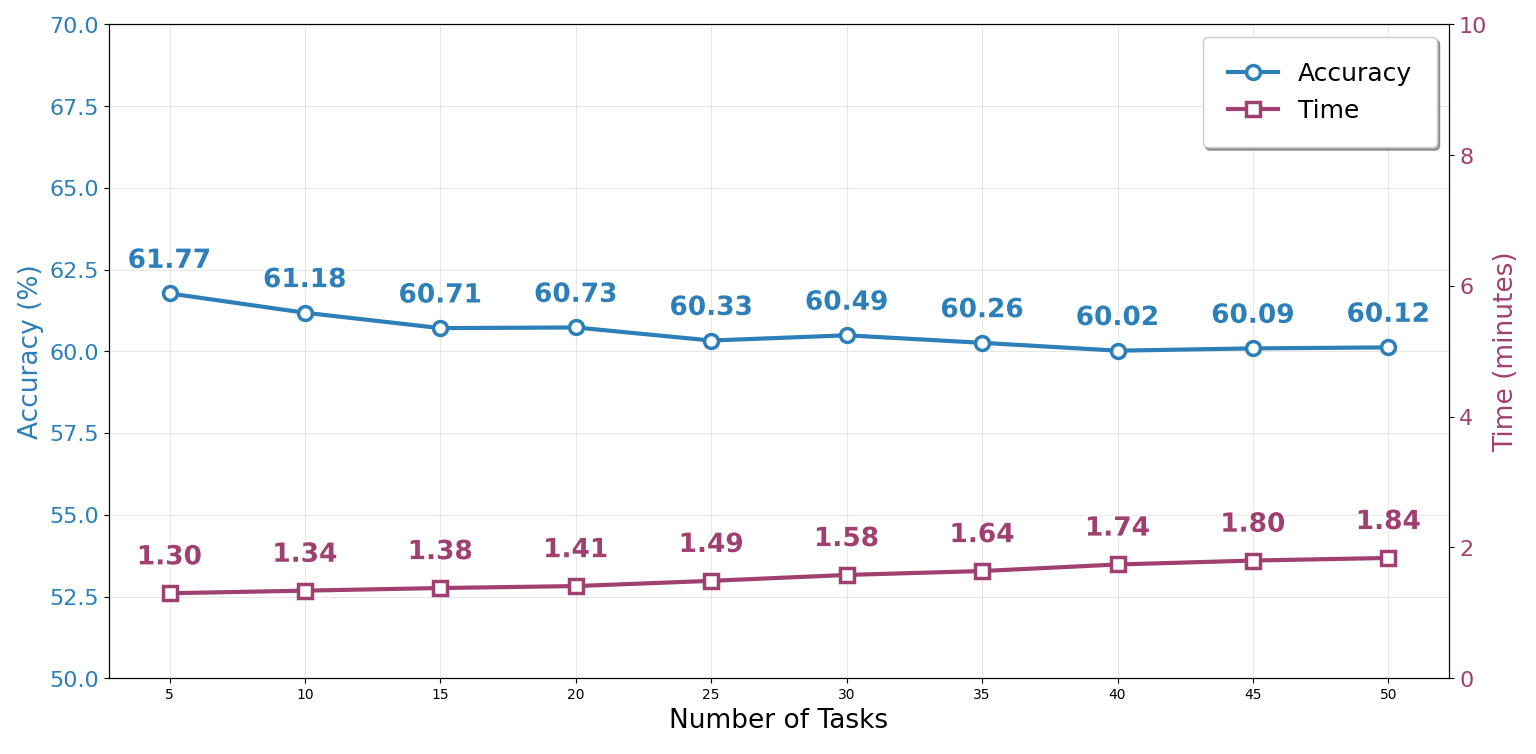}
\caption{HydraCIL ($k=2$) performance measured across different tasks splits on CIFAR-100 dataset.}\label{Fig:incr_task}
\end{figure}

To validate that even as our model creates a new head per task its computational cost scales modestly, we trained HydraCIL using a wide range of task splits for CIFAR-100: from 5 to 50 tasks, using increments of 5. Figure \ref{Fig:incr_task} shows how using 5 heads our model required 1.3 minutes to train, while for training 50 heads required 1.84 minutes. That is an increase of 10x more heads, but only a 42\% increase in training time. For the sake of comparison, notice that increasing the number of tasks by a factor of four (5 to 20 tasks) resulted in a 8\% more training time for HydraCIL, 203\% time for RMM-FOSTER, 359\% for DER, and 9\% for CIFNet.

Table \ref{Tab:gpu_vs_cpu} illustrates the performance of HydraCIL when executed on a CPU compared to a GPU. While the GPU naturally offers accelerated processing, the CPU-based training remains relatively efficient even compared to RMM-FOSTER and DER executed on GPU.


  

\begin{table}[htb!]
    \setlength{\tabcolsep}{4pt}
  \centering
  \caption{Results of HydraCIL trained using a CPU on CIFAR-100.
}\label{Tab:gpu_vs_cpu}
  \begin{tabular}{c|c|c|c|c}
    \hline
    Tasks & k &\begin{tabular}{@{}c@{}}Time \\ (min)\end{tabular} & \begin{tabular}{@{}c@{}}Energy \\  (kWh)\end{tabular} & \begin{tabular}{@{}c@{}}Emissions \\ (kg CO2-eq)\end{tabular} \\
    \hline
    20 & 20 & 28.54 & 0.030 & 0.005 \\
    \hline
     20 & 2 & 28.52 & 0.030 & 0.005 \\
    \hline
     5 & 20 & 28.59 & 0.030 & 0.005 \\
    \hline
     5 & 2 & 28.38 & 0.029 & 0.005 \\

    \hline
  \end{tabular}
  
\end{table}




\section{Conclusions}
\label{Sec:conclusions}

In this work, we propose HydraCIL, a class-incremental learning model that relies on prototypes and multi-head classifiers. Despite its simplicity, it achieves competitive accuracy compared with other approaches across four widely used Class-Incremental Learning datasets: CIFAR-100, ImageNet-100, CoRe50 and Flowers102. Training time and energy consumption are drastically reduced due to the frozen backbone decoupling and isolated lightweight heads training, while maintaining strong performance. Several tests were performed to measure HydraCIL scalability, regarding different numbers of heads, dataset sizes, and number of prototypes ($k$), demonstrating that the approach scales efficiently across a wide range of configurations. Additionally, HydraCIL's performance on CPU surpasses the GPU-accelerated training times of several established baselines. This efficiency, combined with its robust performance, positions HydraCIL as a primary candidate for sustainable, ``Green AI'' applications on edge devices that require rapid adaptation to new domains.

Future lines of research include increasing the complexity of task-specific classifiers, exploring methods to use prototypes directly for prediction, and developing dynamic selection mechanisms for the number of class prototypes.








\begin{footnotesize}

\bibliographystyle{IEEEtran} 
\bibliography{biblio} 



\end{footnotesize}


\end{document}